# Approaches to Artificial General Intelligence: An Analysis

Soumil Rathi                              S̲oumilrathi@gmail.com

**Abstract**

This paper is an analysis of the different methods proposed to achieve AGI, including Human Brain Emulation, AIXI and Integrated Cognitive Architecture. First, the definition of AGI as used in this paper has been defined, and its requirements have been stated. For each proposed method mentioned, the method in question was summarized and its key processes were detailed, showcasing how it functioned. Then, each method listed was analyzed, taking various factors into consideration, such as technological requirements, computational ability, and adequacy to the requirements. It was concluded that while there are various methods to achieve AGI that could work, such as Human Brain Emulation and Integrated Cognitive Architectures, the most promising method to achieve AGI is Integrated Cognitive Architectures. This is because Human Brain Emulation was found to require scanning technologies that will most likely not be available until the 2030s, making it unlikely to be created before then. Moreover, Integrated Cognitive Architectures has reduced computational requirements and a suitable functionality for General Intelligence, making it the most likely way to achieve AGI.

**Keywords:** Artificial General Intelligence, Cognitive Systems, Intelligent Systems, Artificial Intelligence

## 1. Introduction

The evolution of AI over the past few years has been remarkable. AI is being used in fields all across the spectrum, from medicine to education. All of these innovations in Artificial Intelligence are a part of Artificial Narrow Intelligence. This classification can be seen as follows:
1. Artificial narrow intelligence (ANI)- Narrow range of abilities.
2. Artificial general intelligence (AGI)-On par with human capabilities.
3. Artificial superintelligence (ASI)-More capable than a human.

The definitions provided are vague and have been expanded further in the next section, but are useful for an intuitive understanding of the capabilities of future technologies.

AGI could be responsible for conducting most of the menial, everyday tasks conducted today by a human. Hence, the development of AGI is highly likely to aid humankind throughout a lot of fields.

## 2. AGI

For defining AGI, we must first define intelligence. I have used the definition of intelligence as follows: Intelligence is the ability of an object to pick the best action to achieve its goal in any scenario, coupled with the ability of an object to create sub-goals for the eventual achievement of the original goal. Hence, AGI can be defined as an artificial model that can, within any environment, decide the best action to take in order to achieve its goal as well as create sub-goals for the long term achievement of the original goal.

The reasoning behind keeping a goal-oriented approach is that for any AGI, we would wish for it to complete a set of goals given to it by humans. Also, the reasoning behind separating creation of sub-goals and picking the best action has been done to emphasize that many times, the action that seems to be most beneficial to the eventual goal might be less beneficial in the long term, and the AGI should focus on achieving its goal.

Now, AGI can also be defined as a human-level AI, that is, an AI model that can replicate the intelligence within a human. This would bring us to the conclusion that AGI should be able to replicate the human mental processes(Kiely, 2014), which are listed below:
1. Perception
2. Motor control
3. Memory
4. Attention
5. Language
6. Thinking
7. Decision-making

There are many other human cognitive functions which can be mentioned, such as social intelligence, motivation, and reasoning, but the key cognitive processes of the human brain which have been agreed upon are the ones above.

Here, though, we can find a flaw in the terminology of 'human-level'. This is because we will have to make considerations about whether we truly want the AGI to mimic every feature of humans, including negative ones such as biases caused due to emotions and false memories. Moreover, the ability of an object to change its primary goal, though it is an ability humans possess, should not be an ability given to an AGI, because AGI is being built to complete tasks given to it by humans. For the rest of this paper, whenever I refer to the term 'human-level', I will assume the absence of negative, undesirable, features.

### 2.1. AGI compared with narrow AI

Most AI research nowadays can be referred to as "narrow AI", which stands for AI that focuses on solving specific, narrowly constrained problems. A narrow AI program does not 'know' what it is doing, and hence, it cannot generalize what it has learned beyond its narrowly constrained problem domain. In recent examples, a narrow AI program such as AlphaGo(Silver et al, 2016), which is a program having mastered the board game 'Go' cannot transfer those skills into other similar games. Even the more versatile AlphaZero(Silver et al, 2018), which can play 'Shogi' and 'Chess' as well as 'Go' is a narrow-AI program, using reinforcement learning and having played with itself over 5 million times rather than generalization. As a result, these programs showcase excellence in one specialized area, but cannot generalize that skill into another field, even if it is similar to the original one.

AGI is a framework which is in direct contrast with narrow-AI. Artificial General Intelligence is a program that can solve a variety of distinct problems in different fields without the requirement of being re-trained in each field. It should portray a human-like intelligence, and hence be capable of tasks like attention, memory and creativity. A program based on AGI will have the ability to learn the rules of a game like 'Chess', and then generalize that expertise to other similar games such as 'Shogi'.

## 2.2. Consciousness

There has been a long debate about whether or not consciousness plays a role in the creation of AGI. To begin answering this, we must first identify a universal, rough understanding of the word 'consciousness'. In the context of AGI, I am referring to consciousness as the ability to be aware of the stimuli around oneself and use that information to help make future decisions. Under this definition, we can say that AGI will require consciousness because of the following reasons:
1. An AGI agent is expected to make decisions in unknown circumstances. This requires the agent to have a system of automatic input of the environment around it, which refers to being aware of its surroundings.
2. AGI will have to learn a multitude of processes, including actions like walking, speaking, building etc. This requires the agent to have a memory of past observations, allowing it to refer back to its memories to make better observations of the present. Hence, it will need to store its information as memory.

As a result, AGI requires the two main components that comprise consciousness, resulting in a need to implement consciousness into the agent.

## 3. Methods proposed for AGI

There have been multiple proposed methods to reach AGI, which have been shown and analyzed as follows:

## 3.1. Human Brain Emulation

Human Brain Emulation is the idea of simulating the human brain in a computational manner. Models of the human brain are to be created and put together to achieve an artificial brain, which can then be used to power Artificial General Intelligence, as the artificial brain will have similar capabilities to a human brain, such as pattern recognition and adaptability, while also having the advantages of increased computational speed and reliability.

3.1.2. HARDWARE

In 'The Singularity Is Near', Ray Kurzweil(2005) has proposed a human brain emulation solution, stating that AGI will likely be possible by the 2020s-2030s. Kurzweil has divided the challenges toward this goal into hardware and software challenges.

Kurzweil explains that because there are around $10^{16}$ synaptic transactions per second in the brain and $10^3$ calculations per synaptic transaction, the hardware capability required to enable the emulation of the human brain is close to $10^{19}$ calculations per second, but that $10^{16}$ calculations per second is sufficient to achieve functional equivalence of all brain regions, as the models of brain regions require far less computation then is implied by all of the neural components. Kurzweil claimed that we will be able to reach this by 2020. As of writing this, in December 2021, the fastest supercomputer built is the Fugaku supercomputer(Fujitsu, 2021), with a sustained processing speed of 488 petaFLOPS, or around $10^{18}$ calculations per second.

According to Kurzweil's calculations, we have achieved the hardware required to simulate the brain. However, this technology cannot be used in a commonplace or accessible manner yet.

3.1.3. SCANNING

Ray Kurzweil then moves onwards to elaborate that we will need to scan and perform imaging on the brain in order to accumulate data on the precise characteristics and dynamics of the regions of the brain, such as the hippocampus and cerebellum.

He suggests the use of nanobots, which would need to have a radius of 7 to 8 micrometers, to scan brains non-invasively in realtimes, stating that they will be available by the 2020s. As of 2021, nanotechnology is being used for *in vivo* molecular imaging(Sim and Wong, 2021). Taking this into account, it is reasonable to assume that we will be able to manufacture nanobots with a width of 7-8 micrometers for *in vivo* imaging of the human brain within the next decade.

3.1.3. SIMULATING THE BRAIN

Ray Kurzweil believes that 'as the requisite neuron descriptions and brain-interconnection data become available, detailed and implementable replicas will be developed for all brain regions'. He states how once the algorithms for a particular region are understood, they can be refined and extended before being implemented in synthetic neural equivalents, on a computational medium, achieving human brain emulation.

In order to showcase that the mathematical model of the artificial neurons is sufficiently accurate for the human brain, Kurzweil brings up the experiment conducted by the University of California at San Diego's Institute of Nonlinear Science. The experiment involved connecting artificial neurons with real neurons from a lobster in a single network. When a network of neurons receives an input, the signaling amongst them appears to first be frenzied and random. Typically after a fraction of a second, this chaos dies down and a stable pattern of firing signals emerges, which appears to be their decision. When the artificial neurons were added to the network, they mixed well with the real neurons present there and were able to partake in the 'decision making' process.

This method of achieving AGI has seen many projects conducted, one of which being Lloyd Watts' simulation of the auditory cortex(2012).
The simulation has been created by reproducing the transformations performed in the inputs by each region. This model showcases the feasibility of converting neurobiological models into working simulations. Watts' model is capable of features such as the 'cocktail party effect', or choosing one speaker to pay attention to over the rest. In the model, there were drawbacks, such as: the recognition systems such as speaker identification and speech recognition were of the introductory level, and that there was no integration between the different representations, even though the model was able to run them simultaneously. A proper computational model will have to have full integrations between all representations of sound(to be able to extract and use all of the information in the auditory signals), but nevertheless, this model showcased a major achievement in computational neuroscience.

Another development showcases the modeling of the human cerebellum by Hiroshi Yamaura et al(2020). Using the known anatomical structure of the cerebellum(Eccles et al, 1967), the researchers built a neural network replicating it artificially. These researchers reproduced the neurons and the connections between them, having created seven layers in their neural network:

upper molecular layer, lower molecular layer, Purkinje cell layer, granular layer, deep cerebellar nucleus, inferior olive, and pons.

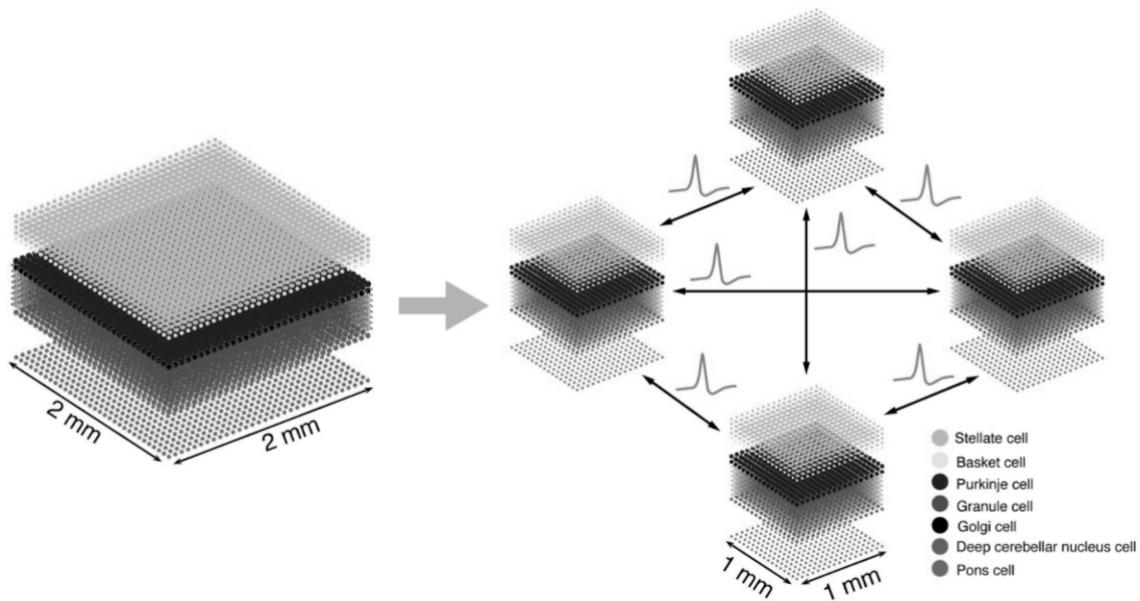

*Figure 1 - This image shows the layers used by Hiroshi Yamaura et al(2020) to represent the cerebellum.*

The model had been tested against a computer simulation of a simple cerebellum-dependent eye movement control task called optokinetic response(OKR), which is an eye movement reflex, which is induced by slow movement of the whole visual field image on the retina. The human cerebellum issues the motor command for eyes to move to the same direction with the visual field movement, so that the blur in the retinal image is reduced. The response by the artificial Purkinje cells, the neurons highly responsible for controlling motor movement, replicated the response shown by real Purkinje cells during animal testing, indicating that the model was able to replicate the functioning of the cerebellum.

These developments showcase that as the technology for the *in vivo* imaging of the brain develops and the human brain is sufficiently scanned, it will be possible to transfer those scans into a computational model replicating those regions.

3.1.4. REVIEW

Ray Kurzweil, in his book, had predicted that we will have achieved sufficient hardware capabilities to run a simulation of the human brain by 2020. Now, in 2021, this prediction has been met. With the rise in the capabilities of in vivo imaging technologies, it seems likely that we will be able to achieve human brain emulation by the 2030s-2040s, thus achieving AGI. The key drawback with development in Human Brain Emulation, however, is that practical efforts to achieve Human Brain Emulation, or even to replicate a section of the brain artificially, requires the development of in vivo technology to scan the section of the brain in question. Another drawback of the human brain emulation system to create AGI is that we cannot allow an artificial model to be created with the full freedom of the human mind. This is because it will allow the AGI to undergo processes such as changing its original goals and prioritizing its own self over anything else, in ways we might not wish. Such drawbacks will arrive when we cannot 'feed' a primary goal into the mind of the AGI, and will have to be studied further.

## 3.2. Algorithmic Probability

The method to create AGI using Algorithmic Probability involves the prediction of the environment around the agent using Solomonoff's induction(2010) and the Bayes theorem, and then, based on the prediction, the agent will choose the action which will maximize future reward upto a certain point, usually taken as the lifetime of the agent. Such a system allows the agent to make the best decision possible in any given situation, without prior knowledge of the environment around it.

### 3.2.1. PREDICTION OF THE ENVIRONMENT

The prediction of the environment occurs on the basis of Ray Solonomoff's induction theorem, which allows the model to come up with a probabilistic measure of the most likely environment, given the observations. Solomonoff's induction works on the idea of a Universal Turing Machine. Intuitively, this is a machine that can take in a binary input of any length, and on the basis of certain commands in the Turing Machine, convert that input into an output, also in binary.

The method for the prediction of the environment is that the observation received by the agent is treated as the output of the Universal Turing Machine, and for all possible inputs into the Turing Machine, the inputs which give the correct output are taken as potential environments of the agent. Once an environment which outputs the desired output, say environment $x$, has been identified, the probability of the environment is judged by the following equation:

$P(x) = 2^{-l(x)}$,

where $P(x)$ is the probability of environment $x$, which has a binary length of $l(x)$.

The prediction of the environment is where the major drawback of algorithmic programming to create AGI is introduced, which is its uncomputability. Since the agent will have to go through an infinite number of potential environments to decide if each one is a solution, the model is uncomputable. Solomonoff has claimed that this is a desired feature, as it represents that the model is the 'ideal' model for taking the best decision possible, and a practical model will have to be created as an approximation of AIXI.

### 3.2.2. AIXI

AIXI is the theoretical model created by Marcus Hutter(2000), involving Solomonoff's induction, to create Artificial General Intelligence. The idea behind the model was to create a model which could take the best action possible in any given, even previously unknown, scenario.

In a hypothetical scenario with infinite computability, AIXI receives two inputs from the environment around it, which is the observations and the rewards. Now, these observations are input into Solonomoff's induction, allowing the model to perceive the model of the environment with highest probability. However, in order to maintain the Epicurus Principle, all of the possible environments are maintained with AIXI, even though the shortest, and hence simplest environment, is given a higher weightage, in the form of probability. This has been represented with the following equation:

$$M(x) = \sum_{p:UTM(p) = x} 2^{-l(p)},$$

where *M(x)* refers to the total probability of observation *x*, and *p:UTM(p)* = x refers to every possible environment *p* for which the observation *x* is possible.

Now, with knowledge of the environment, AIXI computes the best possible action to maximize reward for the next *m* steps, where *m* is an arbitrary number. This can be done by ranking each possible action $a_k$ by the total rewards which will be gained by executing the action.

$$\text{AIXI} \quad a_k := \arg\max_{a_k} \sum_{o_k r_k} \ldots \max_{a_m} \sum_{o_m r_m} [r_k + \ldots + r_m] \sum_{q:U(q,a_1..a_m)=o_1 r_1..o_m r_m} 2^{-\ell(q)}$$

*Figure 2. This image showcases the calculation of value of an action $a_k$ by AIXI, where $r_k$ refers to the rewards gained by the model at step k(Hutter, 2000)*

Intuitively, this equation states that, for a point *k* in its lifetime, for each Turing Machine it uses to predict the environment, if the machine's predictions has been consistent with the observations received, the model will calculate the rewards it will gain based on the prediction of that machine uptil point *m* if it executes action $a_k$. It will then take the action $a_k$ that has the maximum rewards, across all of its Turing Machines.

Because of the uncomputability of AIXI, in order to allow it to possibly function in a practical scenario, approximations of AIXI have been thought of. The two key reasons AIXI is uncomputable are:
1. AIXI must look through an infinite number of potential environments in order to identify the correct environments
2. For some inputs to a Turing Machine, it is possible that the Turing Machine will enter an infinite loop, without halting to output anything.

3.2.3 APPROXIMATIONS OF AIXI

(a) AIXItl -
AIXItl(Hutter, 2000) is a version of the AIXI model that only considers models of environments upto an arbitrary length *l* and limit the computation time of the Turing Machine upto time *t*. Such an approximation, while it does make AIXI computable, still suffers from the same problem. The computation time of AIXItl is $t \cdot 2^l$, which is impossible to reasonably compute. This is because, to achieve an approximation of AIXI that can replicate the original AIXI, $l \rightarrow \infty$. This would also mean that the computation time of AIXItl would slowly reach infinity. Even to simulate the environment around a human, the length l would be too large for $2^l$ to be practically possible to compute.

(b) UCAI -
Susumu Katayama(2019) had taken on the goal to create an approximation of AIXI that is stronger than AIXItl, calling it Unlimited Computable AI(UCAI). This approximation attempts to solve the two problems faced by AIXI in a different manner. In the place of a Turing complete language to run the Universal Turing Machine in AIXI, Katayama has chosen to use typed lambda calculi, which is a terminating language. This change automatically makes the model Turing-incomplete, as it cannot perfectly replicate a Turing Machine, since the language is a terminating one. However, the author has claimed that this would not impact the implementation, as the model can still replicate the functionality of AIXItl with timeout *t*. To solve the problem of infinite models, Katayama has suggested the use of an infinite stream to represent the possible environments. This has been done to allow the stopping of the stream at an arbitrary point, which he claims is a stronger method than limiting the length of the input, as has been done in AIXItl.

While this method may be stronger than AIXItl, it suffers from the same issues, being computation time. To simulate the environment around a human, the arbitrary point to stop the infinite stream, say $x$, will have to be a large value, such that the computation time with factor $2^x$ would not be practically possible.

### 3.2.4. REVIEW

The theory behind AIXI is sound, and the model can well be used as a benchmark for what AGI must be capable of. However, so far, the approximations that have been made of AIXI are not fast enough to be used in the practical scenario. Until an approximation is developed that allows environments with a high binary length $l$ to be computed, something which seems unlikely given current technological progress, it will not be possible for an AIXI approximation to achieve AGI.

## 3.3. Integrative Cognitive Architecture

The idea of an Integrative Cognitive Architecture to create AGI is to identify the central cognitive processes that occur within the human brain and then replicate them individually within an AGI. The software has been introduced by Ben Goertzel et al with the idea of CogPrime(2014).

### 3.3.1. COGNITIVE ARCHITECTURES

A cognitive architecture is a theory about how the human mind is structured. These determine what the fixed mechanisms and processes that an intelligent agent will use to complete tasks are. There are 3 types of cognitive architectures:

Symbolic - Knowledge is represented in the form of symbols, much like the way a computer processes input and output. For example, in the sentence "The apple is red ", the symbol of apple can be seen as an object that is red, and the symbol "red" can be seen as a representation of the color of the apple. This mechanism focuses on "working" memory which draws on long term memory for inferences, and maintains a centralized control over perception, cognition and action. In practice, such architectures tend to be weak in learning, creativity, episodic and associative memory, all of which are key to achieve AGI(Goertzel et al, 2014). One project undertaken in this area is SOAR(Laire, 2012).

Subsymbolic/Emergent - In an emergent system, the cognitive architecture can be compared to the functioning of a neural network. Like a small collection of neurons in the brain, a sub-symbolic system is composed of a small collection of perceptrons - linear binary classifiers - that operate in parallel to recognize a given input(Asselman et al, 2015). This recognition process is accomplished by the adjustment of the weights which connect the perceptrons to each other. A collection of nodes can thus be enabled to recognize a given input and produce a specified output by adjusting the weights of the connections between the perceptrons. So far, it has not been possible to achieve abstract reasoning or complex language processing using simply an emergent approach(Goertzel et al, 2014). Until giving rise to those is understood, it is not possible for AGI to be created only with subsymbolic approaches. DeSTIN(Arel and Coop, 2009) is one project being undertaken with a subsymbolic cognitive architecture.

Hybrid - These architectures typically function as a mix of both symbolic and subsymbolic architectures, both of which would supplement each other. This combination allows the symbolic agents to manage reasoning and understanding of language, while the subsymbolic agents would be responsible for learning and creativity. However, it must be taken into account that these agents would need to be synergic with each other, in order to maximize functionality as well as computational ability.

CogPrime functions as a hybrid architecture. Goertzel has made sure to emphasize the importance of maintaining cognitive synergy between this hybrid architecture, detailing that if cognitive synergy is not present, the computation time will be too high to be reasonable. This has been expanded on in **Section 3.3.5**.

### 3.3.2. WORKING OF THE COGPRIME SYSTEM

CogPrime is intended to take the best action possible at every step in a previously unknown environment to help achieve a goal. The system does so by defining a central process:

$$Context\ +\ Procedure\ \rightarrow Goal\ <p>$$

This indicates that, while context *C* holds, we conduct procedure(action) *P* to achieve goal *G* with certainty *p*.

This allows for two major functions for CogPrime, which are analysis and synthesis. Analysis refers to analyzing the probability *p* of achieving goal G, and synthesis refers to filling in one or more of the missing elements in the process. **Section 3.3.3.** and **Section 3.3.4.** explains briefly how synthesis is conducted within CogPrime, in order to analyze the functionality of such a model.

### 3.3.3. IDENTIFYING THE CONTEXT

Identification of the context is done by deductive inference, conducted by probabilistic logical networks(Goertzel et al, 2014).

Synthesis of the context is conducted with reference to the declarative memory stored in CogPrime, which refers to the memory of knowing certain facts such as London is the capital of English and that dogs are animals. This memory is under a constant stream of inference, for example, if the agent knows the following two facts: Cats are mammals; mammals have fur, it will be able to infer that cats have fur, which it can then use in any relevant situation. This is always done in a probabilistic formula, in order to reasonably calculate confidence levels, and store multiple hypotheses.

### 3.3.4. IDENTIFYING AND SELECTING PROCEDURES

Similar to the synthesis of the context by the PLN, the synthesis of the procedures to be carried out is best conducted by an algorithm known as MOSES(Goertzel et al, 2014). MOSES is the cognitive process associated with procedural memory, which refers to memory about actions which the agent can conduct, to avail different rewards.

Procedural synthesis allows us to select the best procedure for the achievement of a specific goal under any specific context, by deducing the effectiveness of the action on the bases of past experiences.

### 3.3.5. COGNITIVE SYNERGY

To enable an effective learning model for the agent, it is important that all the cognitive processes work together. In case that does not happen, it will lead to slower computations, and in many cases, no good answers. For example, it is possible that any cognitive process, say MOSES, gets 'stuck' in one of its computations, unable to produce a high-confidence output. Without cognitive synergy, the agent will not be able to achieve a high-confidence output, as has been desired.

However, if the agent has cognitive synergy, MOSES can refer to other processes such as PLN, utilizing the Declarative memory to reach an answer. While many people believe that it might be better to have all memory and processes stored as one, instead of splitting, this has high potential to lead to Combinatorial Explosion, caused by a rapid increase in input size and dimensions. The utilization of synergic memory and processes will allow these processes to avoid that.

3.3.6. REVIEW

Ben Goertzel's CogPrime agent simulates the primary processes of a human brain through the replication of various cognitive processes, such as inference and attention allocation. This allows the agent to perform accurate context selection, followed by action selection, choosing the best action for long term benefit.

This model introduced by OpenCog is similar to the AIXI model described by Marcus Hutter, as shown in the previous section. While AIXI relied on Solomonoff's induction to identify the current situation of the agent, CogPrime relies on an inference method based on PLNs. This allows it to overcome the primary disadvantage of AIXI and its approximation, being their uncomputability. In terms of the design of the model, it roughly replicates the functionality of AIXI, which has been mathematically proven to be a functioning model(theoretically) for AGI.

As a result, CogPrime is the most promising route to reach AGI. While Kurzeil's Human Brain Emulation would function as AGI, implementation of such a project requires intense computational resources as well as *in vivo* technology that has not been developed yet, making it impossible to achieve under current scenario.

## 4. Conclusion

In this work, I primarily analyzed the different models and projects being undertaken. So, as we can see from these models, there is not one single correct answer to build AGI. There are multiple possible methods, but they all have separate timelines and drawbacks.

Human Brain Emulation is a theory that has the potential to perfectly replicate the human brain computationally, and thus create a successful AGI program. The drawbacks of this method are the intense computational capabilities required - $10^{16}$ computations per second - which, as of the time of this paper, is possible only in rare and expensive supercomputers. Moreover, it also requires *in vivo* scanning technology that will not be created until the 2030s - 2040s.

OpenCog's CogPrime system is a model that mimics the cognitive functions of a brain to identify information about its surroundings, and uses an action-selection module to identify the best action to take in a scenario. This allows it to create an intelligent model, by the definition provided in the beginning of the paper, and hence create an AGI program. The drawbacks of this model are that the successful separation of memory types will have to be undertaken, alongside a synergy between all of them, in order to create a computationally effective system. Overall, the process that will yield the quickest and most efficient results is CogPrime, because the technologies required for the creation of CogPrime are mostly available and its computational ability is much below that of a computational human brain.

Approaches to Artificial General Intelligence: An Analysis